\documentclass{llncs}
\usepackage{makeidx}
\usepackage{graphicx}
\usepackage{multirow}
\usepackage{tabularx}
\usepackage{CJKutf8}
\usepackage{caption}
\usepackage{graphicx}
\usepackage{subfigure}
\usepackage{booktabs}
\begin{document}
\frontmatter          
\pagestyle{headings}  

\title{Machine Learned Resume-Job Matching Solution}

\author{Yiou Lin \and Hang Lei \and Prince Clement Addo \and Xiaoyu Li}

\authorrunning{Yiou Lin et al.} 
\institute{University of Electronic Science and Technology of \\China, 
Chengdu, 610054, China\\
\email{lyoshiwo@gmail.com   hlei@uestc.edu.cn pricetheboy@gmail.com xiaoyuuestcg@uestc.edu.cn} }

\maketitle              
\makeatletter

\makeatother

\begin{abstract}
Job search through online matching engines nowadays are very prominent and beneficial to both job seekers and employers. But the solutions of traditional engines without understanding the semantic meanings of different resumes have not kept pace with the incredible changes in machine learning techniques and computing capability. These solutions are usually driven by manual rules and predefined weights of keywords which lead to an inefficient and frustrating search experience. To this end, we present a machine learned solution with rich features and deep learning methods. Our solution includes three configurable modules that can be plugged with little restrictions. Namely, unsupervised feature extraction, base classifiers training and ensemble method learning. In our solution, rather than using manual rules, machine learned methods to automatically detect the semantic similarity of positions are proposed. Then four competitive ``shallow" estimators and ``deep" estimators are selected. Finally, ensemble methods to bag these estimators and aggregate their individual predictions to form a final prediction are verified. Experimental results of over 47 thousand resumes show that our solution can significantly improve the predication precision current position, salary, educational background and company scale.  

\end{abstract}
\section{Introduction}
Following the effects of the global financial crisis in 2008, large financial institutions have collapsed. Knowledge workers, even in wealthiest countries have to worry about losing their well-paid, full-time jobs, and cannot easily find similar ones elsewhere. An effective e-recruiting engine can help job seekers easily access recruitment opportunities and reduces the recruitment labor by providing suitable items which match their personal interests and qualifications. It also frees companies from information overload and advertisement cost. The key module for a dynamic e-recruiting engine is job matching system which makes an effort to engage the unemployed who are well suited to the vacancies to be filled. 

In this work, we evaluate the job matching problem as a classification problem. This is to identify a job seeker's current employment detail (the last position in the resume) by their previous employment history. The framework of our solution is constructed by several modules based on keras\cite{chollet2015keras} and sklearn\cite{scikit-learn}, and can be practically deployed and easily verified. Through empirical evaluation, we show step by step how to intensify the solution and get a better performance than a baseline manual rule-based job matching system.

The rest of the paper is organized as follows. In Section 2 we survey the related literature to give an overview of the research background. In Section 3, we introduce the dataset description. Then we propose the feature extract methods and the machine learned models in Sections 4 and 5 respectively. In Section 6, we report two ensemble methods and analyze the empirical results. Finally, the paper is concluded in Section 7 with our future work.

\section{Literature Review}
Job matching system is a kind of recommender system. Recommender system was first introduced by Resnick and Varian \cite{Resnick1997Recommender} who pointed out that in a typical recommender system people provide recommendations as inputs, which the system then aggregates and directs to appropriate recipients. After that recommender systems are being highly accepted in various industries and academic areas and are gaining momentum over the years. In general, recommender systems are applied in various domains (such as books, digital products, movies, music, TV programs, and web sites) and help users to find content, products, or services by aggregating and analyzing suggestions and behaviors from other users\cite{Carrer2012Social}\cite{Lu2015Recommender}. In a detailed survey paper\cite{Lu2015Recommender} provided researchers with the state-of-the-art knowledge on recommender system including real-world applications, recommendation methods, real-world application domains and application platforms.

For job matching area, many researches have been conducted to discuss different recommender system related to the recruiting problem as well\cite{al2012survey}. Among them, Malinowski et al.\cite{Malinowski2006Matching} discussed a bilateral matching recommendation systems to bring people together with jobs using an Expectation Maximization (EM) algorithm, while Golec and Kahya\cite{Golec2007} delineated a fuzzy model for competency-based employee evaluation and selection with fuzzy rules. Paparrizos et al.\cite{paparrizos2011machine} used Decision Table/Naive Bayes (DTNB) as hybrid classifier. Though these system used many manual attributes and various information retrieval techniques, compared to our work which employed deep learning methods to accelerate the process of finding the most appropriate jobs, they still failed in keeping with the rapid changes in computing capability and machine intelligence. Similar to the work of Zhang et al.\cite{Zhang2015P}, our work also tries to optimize the knowledge worker-position matching, considering various characteristics of knowledge workers. Compared to the work of Guo et al.\cite{Guo2016R}, Our solution is completely data-driven, without using exterior semantic tool (NLTK and DBpedia) like they did. Besides, our resume data are widely collected from various areas which makes our solution more universal and robust.

\section{Dataset description}
\begin{figure}[h]
\centering
\includegraphics[width=0.9\textwidth]{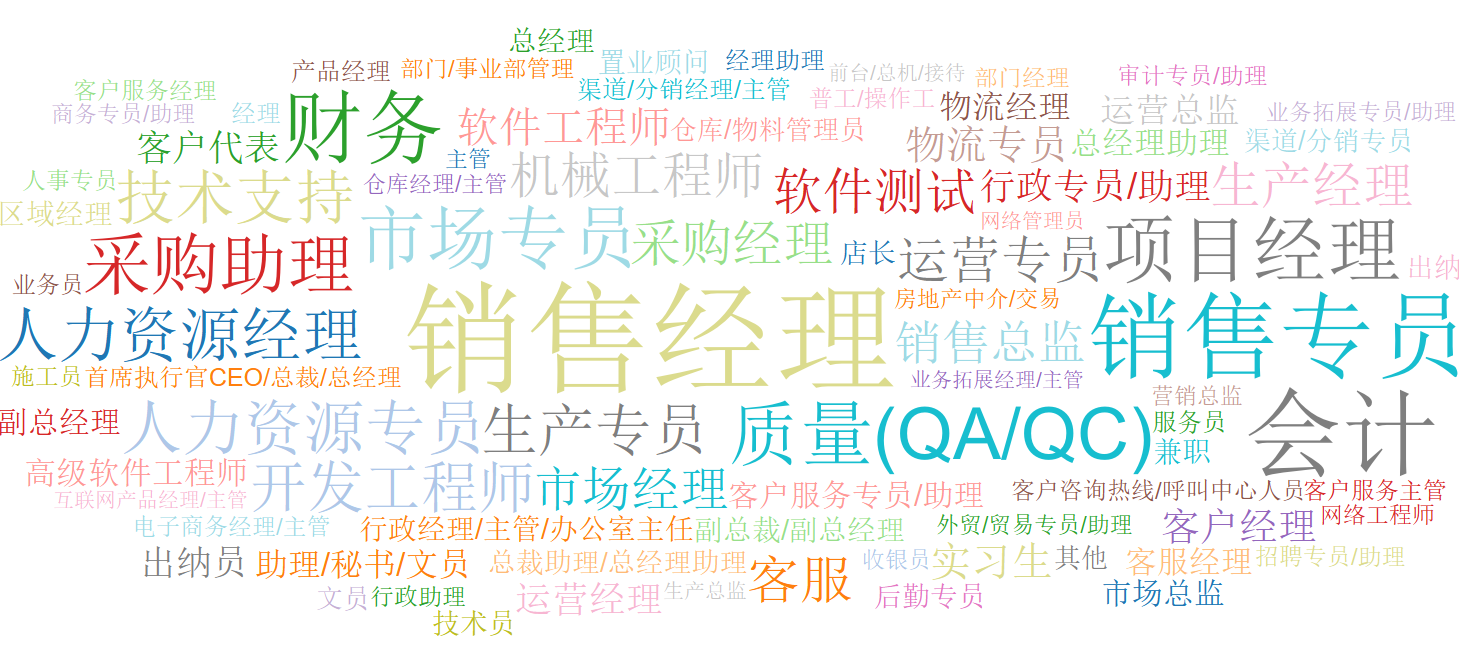}
\caption{Most Frequent Positions which Appear in the Dataset}
\label{fig:position}
\end{figure}

The dataset used was tapped from a job recommend game\footnote{http://www.pkbigdata.com/common/cmpt\_list/all\_all\_time\_1.html} and can be freely downloaded\footnote{http://www.pkbigdata.com/common/cmptData/147.html}. The original dataset contains 70,000 resumes with 34,090 different positions. After cleaning and filtering, 47,346 resumes whose last jobs belong to a particular predication list of most frequent 32 positions (e.g. software engineer, cashier and project manager) were used. Even though, there are 18,736 different positions in the dataset. The most frequent positions are shown in Figure \ref{fig:position}.

Table \ref{table:resume} is an example of resume with job seeker's personal information and work history. 
In particular, salary $ \in \left[ {0,6} \right]$, degree$ \in \left[ {0,2} \right]$ while size means the company's scale. Our problem description can be shown as identifying a job seeker's current position (the last position in the resume) by their previous employment history.

\begin{CJK}{UTF8}{gbsn}

\begin{table}[h]
\renewcommand{\arraystretch}{1.0}
\centering
\caption{A Job Seeker's Resume}
 \begin{tabular*}{0.95\columnwidth}{@{\extracolsep{\fill}}lllll}
  \toprule
  \multicolumn{2}{c}{The Json Stucture}&\multicolumn{2}{c}{One of the workExperienceList }\\
  \midrule
  id&558d...761&size&3\\
  major & 通信工程 & salary &	4  \\
degree&1&end\_date& 今  \\
\_id&Object\{...\}&start\_date&2014-8\\
gender&男&industry&计算机/互联网\\
age&31&position\_name&软件测试\\
wokExperimentList&Array[3]&department&硬件测试\\

  \bottomrule
  \\
 \end{tabular*}
 \label{table:resume}
  \end{table}
\end{CJK}

\section{Information Extraction from Resumes}
There are three main kinds of features made up of 95 foundational manual features, 72 cluster features and 380 semantic features. In all, there are 551 features for each resume.
\subsection{Manual Features}
In the beginning, a lot of manual features were designed. Some features are character type, some boolean values, others are numerical. For the example shown in Table \ref{table:resume}, the features include but not limited to gender, age, major, the details and changes of previous jobs,  the age when first employed, the highest salary and so on. The values of character features were inputted into a dictionary, and numerical keys were used instead of the values when training.
\subsection{Cluster Features}
The design features can be transformed from characters to numerical keys, but similar phrases (such as software engineer and Android engineer) would never be identified. Thus, we train a Chinese Word2Vec model using all the resumes' text. And the average word embeddings of a phrases now can represent its semantic meaning. A simple K-mean method was used to classify phrases into 64 and 128 clusters. The last 5 job experiences in a resume were kept, each job experience consists of 7 phrases, including department, industry, position name, salary, size, type (is empty usually) and quarter (if a seeker works 4 quarters for a company will be marked as ``quarter\_4''). Some clusters are shown in Table \ref{table:cluster}.\begin{CJK}{UTF8}{gbsn}

\begin{table}[h]
\renewcommand{\arraystretch}{1.0}
\centering
\caption{Phrases in Different Clusters}
 \begin{tabular*}{0.95\columnwidth}{@{\extracolsep{\fill}}lclclclcl}
 
  \toprule
  cluster0&&cluster1&&cluster2&&cluster3\\
  \midrule
商务部门经理 &&钳工/钣工&&营运/人力资源 &&供应链管理 \\ 
工程经理  &&机械/模具设计/制造&&人事科员 && 采购/外贸&   \\
信审经理 &&工地电工助理&&行政人事办主任 &&  销售渠道 \\
高速项目经理 &&建筑环境/设备工程 &&培训-培训策划  &&跟单员/品质专员  \\
  \bottomrule
  \\
 \end{tabular*}
 \label{table:cluster}
  \end{table}
\end{CJK} 
The rest cluster features are document features. We use LDA to classify resumes into 32 and 64 topics respectively. In all, there are 72 cluster features.
\subsection{Semantic Features}
Differently from cluster features, in this we try to find the semantic meaning of phrases and the potential relationship of the employment history directly. According to the common sense in NLP, the meaning of a word is decided by its context and similar words have similar contexts. Thus, we transformed the work experiences of a resume into a ordered list of (n*7+3) phrases (n experiences in a resume with 3 extra phrases including age, major and gender). Assume each Chinese phrase as a word and each list of phrases as a sentence, after word2vec training, each Chinese phrase would be presented by a vector of 10 dimension and an example of similar semantic meaning is shown in 
 Table \ref{table:semantic}.
\begin{CJK}{UTF8}{gbsn}

\begin{table}[h]
\renewcommand{\arraystretch}{1.0}
\centering
\caption{ An Example of Similar Semantic Meaning}
 \begin{tabular*}{0.95\columnwidth}{@{\extracolsep{\fill}}lclclcl}
  \toprule
  &similarity&&&&similarity&\\
  \midrule
  软件工程师 & .938 & 高级软件工程师 &	&电子商务 & .900 & 艺术设计 \\
软件工程师&.932&开发工程师& &	电子商务 & .876 & 国际经济/贸易 \\
软件工程师&.890&软件测试&	&电子商务&.864&英语\\
软件工程师&.864&技术支持&	&电子商务&.844&广告学\\

  \bottomrule
  \\
 \end{tabular*}
 \label{table:semantic}
  \end{table}
\end{CJK}

\section{Machine Learned Models}
\subsection{Shallow Estimators}
\begin{figure} 
\centering 
\subfigure[] { \label{fig:a}     
\includegraphics[width=0.45\columnwidth]{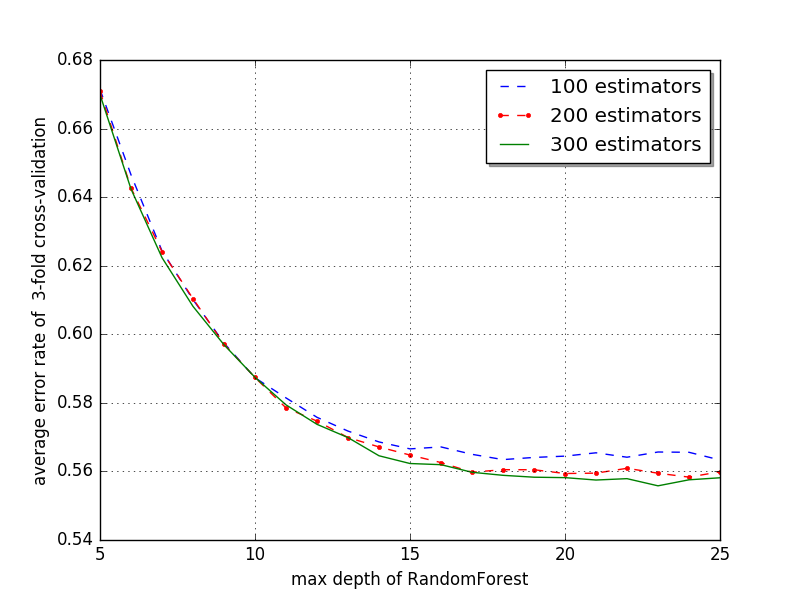}  
}     
\subfigure[] { \label{fig:b}     
\includegraphics[width=0.45\columnwidth]{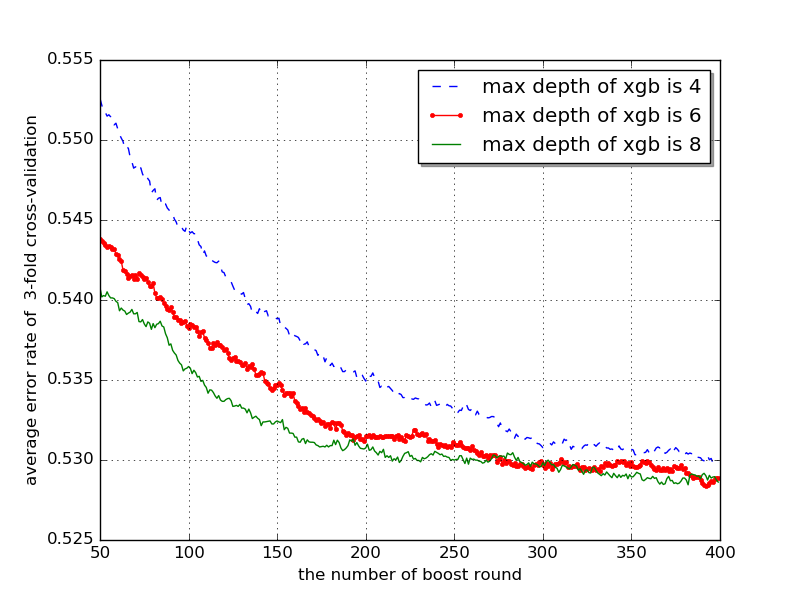}     
}       
\caption{ The Performance of RF and XGB with Different Parameters}     
\label{fig:rfxgb}     
\end{figure}
 Random Forests (RF) is an ensemble learning method using the general technique of random decision trees. Each tree in the ensemble is built from a sample drawn from the training set and the best split is picked among a random subset of the features. XGBoost (XGB), a short form for "Extreme Gradient Boosting", is an optimized distributed gradient boosting library designed to be highly efficient, flexible and portable. The grid search results of Results of RF and XGB are shown in Figure \ref{fig:rfxgb}. After analysis, we find that 473 features are used in XGB while only 163 features are selected by RF.
\subsection{Deep Estimators}
\begin{figure} 
\centering    
\includegraphics[width=0.9\textwidth]{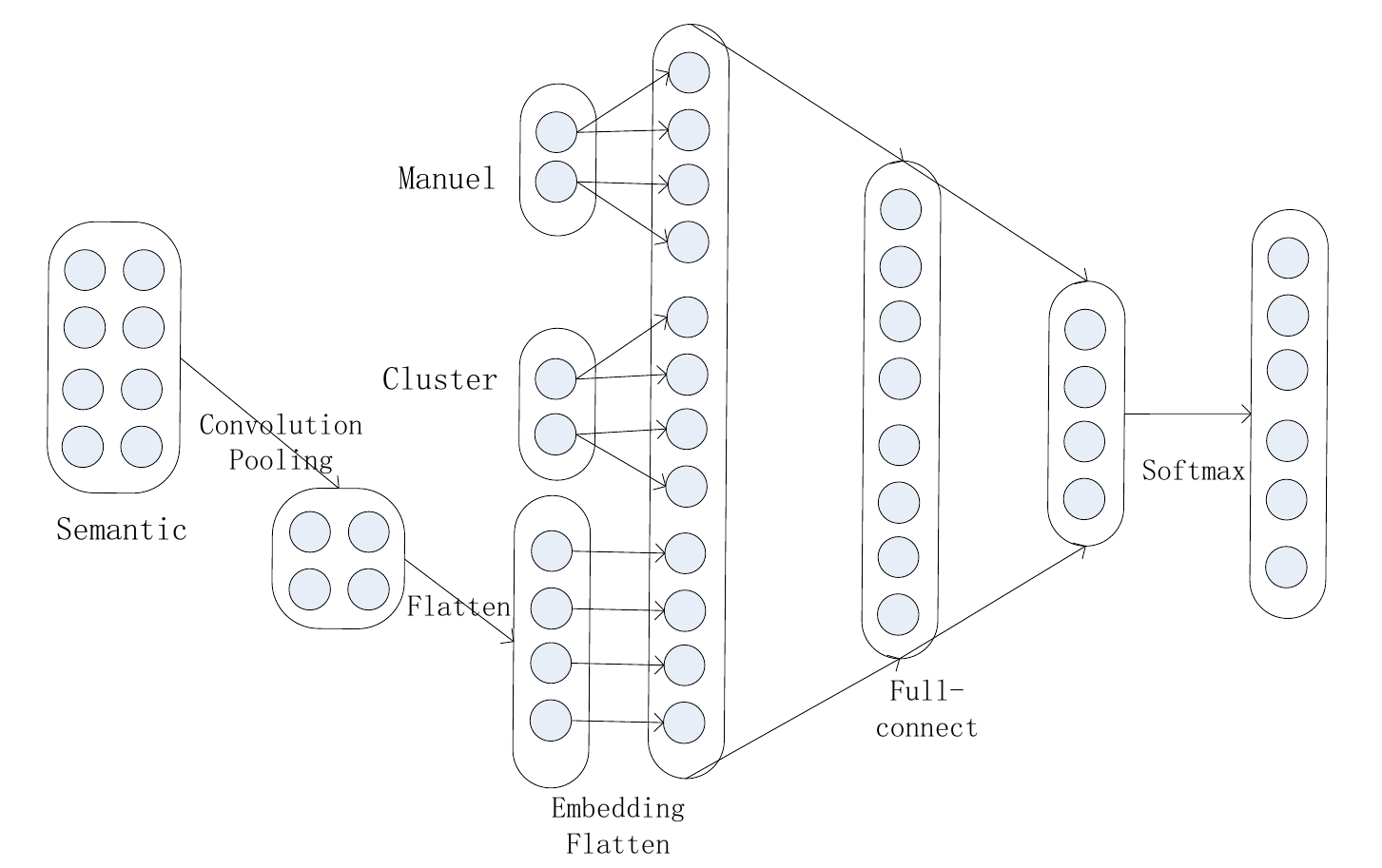}
\caption{The Architecture of CNN Model}  
\label{fig:cnn}
\end{figure}

LSTM is a recurrent neural network which is well-suited to learn from experience to classify, process and predict time series. CNN is formed by a stack of distinct layers that transform the input volume into an output volume through Convolutional layers and Pooling layers. The architecture of our CNN is as shown in Figure \ref{fig:cnn}.

\section{Ensemble Method and Results Analysis}
Bagging is one of the earliest and simplest ensemble based algorithms. Usually, individual classifiers will be combined by taking a simple majority vote of their decisions. Assume that there are three classifiers to make a positive or negative predication. We improve the bagging method (named IBagging) by voting according to the sum of decision probabilities and can easily be extended to multi-class ensemble.
Without any information retrieval techniques and machine learning methods, the basal manual rule will recommend the most frequent label as the recommend item. Then, we can measure our resume-job matching solution in two ways. One is precision, whose goal is to cover as many of correct positions. The results are as shown in Table \ref{table:results}. By analyzing the experiments, we can find that XGB performs best among four basal estimators with longest training time, while CNN model convergences in shortest time with acceptable precision. In the meantime, our solution benefits from both bagging methods and our semantic unsupervised feature extract method.
\begin{table}[h]
\renewcommand{\arraystretch}{1.0}
\centering
\caption{The Precision of Different Classifiers}
 \begin{tabular*}{0.95\columnwidth}{@{\extracolsep{\fill}}llccccl}
  \toprule
   &based on&degree& salary&size &position&training time\\
  \midrule

XGB& manual features&.676&.509&.392&.460&20m 6s \\
XGB& semantic features&.685&.498&.391&.458&41m 24s \\

XGB& all features&.704&.511&.396&.467&53m 19s\\

RF& all features&.666&.511&.394&.453&7m 58s\\

CNN& all features&.695&.508&.391&.465&\textbf{1m 14s}\\
LSTM& all features&.696&.507&.390&.454&5m 52s\\

Bagging& all features&.699&\textbf{.517}&.396&.476&-\\
IBagging & all features&\textbf{.710}&.516&\textbf{.397}&\textbf{.477}&-\\
Manual Rule & frequent item&.484&.254&.256&.141&-\\

  \bottomrule
  \\
 \end{tabular*}
 \label{table:results}
 \end{table}

As we know, lots of resume-position pairs may not appear in testing data, but they are reasonable and often quit similar to those correct pairs over the training dataset. Thus, the other evaluation method, recall for Top-$N$ recommendations is used to evaluate different matching solutions. In this case, recall is the proportion of the correct position from the testing dataset. There are 32 possible position for a resume, with their probabilities given by classifiers, the solution recommends top $N$ positions to a given resume, and reports recall for various values of $N$. The results of Top-$N$ are as shown in Table \ref{table:topn}. The results shows a significant improvement in recall for Top-$N$ using IBagging method compared to the baseline method. 
\begin{table}[h]
\renewcommand{\arraystretch}{1.0}
\centering
\caption{Recall for Top-$N$ Recommendations}
 \begin{tabular*}{0.95\columnwidth}{@{\extracolsep{\fill}}lcccccc}
  \toprule
   &&Manual Rule&&&IBagging&\\
   &$N=2$&$N=3$&$N=4$&$N=2$&$N=3$&$N=4$\\
  \midrule
size&.467&.664&.784& .629&.783&.898\\

degree&.929&1.00&1.00& .965&1.00&1.00\\

salary&.394&.573&.712& .800&.920&.971\\

position&.223&.299&.373& .647&.726&.780\\

  \bottomrule
  \\
 \end{tabular*}
 \label{table:topn}
 \end{table}
\section{Conclusion and Feature Work}
In this paper, we have considered the resume-job matching problem and proposed a solution by using unsupervised feature extraction, surprised machine learning methods and ensemble methods. Our solution is completely date-driven and can detect similar position without extra semantic tools. Besides, our solution is modularized and can rapidly run on GPU or simultaneously run on CPU. Compared to a manual rule-based solution, our method shows better performance in both precision and Top-$N$ recall. Our code is now public and can be tapped from Github\footnote{https://github.com/lyoshiwo/resume\_job\_matching}.
In the future, with more information to be snatched from website, our solution could be extended by including location information, professional skills and description of requirements from both job seekers and employers.
\section{Acknowledge}
This work is supported by the National Science Foundation of China (Grant Nos. 61502082) and the Fundamental Research Funds for the Central Universities (ZYGX2014J065).

\bibliography{jobMatch}

\begin{thebibliography}{10}
\providecommand{\url}[1]{\texttt{#1}}
\providecommand{\urlprefix}{URL }

\bibitem{al2012survey}
Al-Otaibi, S.T., Ykhlef, M.: A survey of job recommender systems. International
  Journal of the Physical Sciences  7(29),  5127--5142 (2012)

\bibitem{Carrer2012Social}
Carrer-Neto, W., Hernández-Alcaraz, M.L., Valencia-García, R.,
  García-Sánchez, F.: Social knowledge-based recommender system. application
  to the movies domain. Expert Systems with Applications  39(12),  10990--11000
  (2012)

\bibitem{chollet2015keras}
Chollet, F.: Keras. \url{https://github.com/fchollet/keras} (2015)

\bibitem{Golec2007}
Golec, A., Kahya, E.: A fuzzy model for competency-based employee evaluation
  and selection. Computers \& Industrial Engineering  52(1),  143--161 (2007)

\bibitem{Guo2016R}
Guo, S., Alamudun, F., Hammond, T.: Résumatcher: A personalized résumé-job
  matching system. Expert Systems with Applications  60,  169--182 (2016)

\bibitem{Lu2015Recommender}
Lu, J., Wu, D., Mao, M., Wang, W., Zhang, G.: Recommender system application
  developments: A survey. Decision Support Systems  74(C),  12--32 (2015)

\bibitem{Malinowski2006Matching}
Malinowski, J., Keim, T., Wendt, O., Weitzel, T.: Matching people and jobs: A
  bilateral recommendation approach. In: null. p. 137c (2006)

\bibitem{paparrizos2011machine}
Paparrizos, I., Cambazoglu, B.B., Gionis, A.: Machine learned job
  recommendation. In: Proceedings of the fifth ACM Conference on Recommender
  Systems. pp. 325--328. ACM (2011)

\bibitem{scikit-learn}
Pedregosa, F., Varoquaux, G., Gramfort, A., Michel, V., Thirion, B., Grisel,
  O., Blondel, M., Prettenhofer, P., Weiss, R., Dubourg, V., Vanderplas, J.,
  Passos, A., Cournapeau, D., Brucher, M., Perrot, M., Duchesnay, E.:
  Scikit-learn: Machine learning in {P}ython. Journal of Machine Learning
  Research  12,  2825--2830 (2011)

\bibitem{Resnick1997Recommender}
Resnick, P., Varian, H.R.: Recommender systems. Communications of the Acm
  40(3),  56--58 (1997)

\bibitem{Zhang2015P}
Zhang, L., Fei, W., Wang, L.: P-j matching model of knowledge workers. Procedia
  Computer Science  60(1),  1128--1137 (2015)

\end{thebibliography}
\end{document}